\documentclass[letterpaper, 10 pt, conference]{ieeeconf}  

\IEEEoverridecommandlockouts    

\usepackage{bicaption}  
\usepackage{graphicx}   
\usepackage{array}      
\usepackage{lipsum}
\usepackage{booktabs}   
\usepackage{amsmath}    
\usepackage{tabularx}

\title{\LARGE \bf
SCAFusion: A Multimodal 3D Detection Framework for Small Object Detection in Lunar Surface Exploration
}

\author{Xin Chen$^{1}$, Kang Luo$^{1}$, Yangyi Xiao$^{2}$, Hesheng Wang*
\thanks{$^{1,2}$Department of Automation, Key Laboratory of System Control and Information Processing of Ministry of Education, Key Laboratory of Marine Intelligent Equipment and System of Ministry of Education, Shanghai Engineering Research Center of Intelligent Control and Management, Shanghai Jiao Tong University, Shanghai 200240, China.
        {\tt\small pinocchio@sjtu.edu.cn, luokang25@sjtu.edu.cn}}%
\thanks{$^{*}$Corresponding author email: wanghesheng@sjtu.edu.cn}
}

\begin{document}

\maketitle
\thispagestyle{empty}
\pagestyle{empty}

\begin{abstract}
Reliable and precise detection of small and irregular objects, such as meteor fragments and rocks, is critical for autonomous navigation and operation in lunar surface exploration. Existing multimodal 3D perception methods designed for terrestrial autonomous driving often underperform in off-world environments due to poor feature alignment, limited multimodal synergy, and weak small‑object detection. This paper presents SCAFusion, a multimodal 3D object detection model tailored for lunar robotic missions. Built upon the BEVFusion framework, SCAFusion integrates a Cognitive Adapter for efficient camera‑backbone tuning, a Contrastive Alignment Module to enhance camera‑LiDAR feature consistency, a Camera Auxiliary Training Branch to strengthen visual representation, and—most importantly—a Section-aware Coordinate Attention mechanism explicitly designed to boost the detection performance of small, irregular targets. With negligible increase in parameters and computation, our model achieves 69.7\% mAP and 72.1\% NDS on the nuScenes validation set, improving the baseline by 5.0\% and 2.7\%, respectively. In simulated lunar environments built on Isaac Sim, SCAFusion achieves 90.93\% mAP, outperforming the baseline by 11.5\%, with notable gains in detecting small meteor‑like obstacles.

\end{abstract}

\section{Introduction}
\label{sec:intro}

Autonomous robotic systems for lunar exploration are required to perceive and navigate through highly unstructured, rugged terrains, which are often populated with small, irregular obstacles such as meteor fragments, boulders, and rocky debris. Reliable detection of these small-scale objects is critical for essential robotic functions including path planning, hazard avoidance, and overall mission safety. Although Bird’s-Eye-View (BEV) based multimodal fusion has established itself as a standard paradigm in terrestrial autonomous driving, its direct application to lunar operational scenarios remains constrained due to several persistent limitations. These include poor detection performance on small and distant targets that are prevalent in lunar environments, misalignment between geometric features from LiDAR and semantic features from cameras which undermines fusion efficacy, the computationally prohibitive cost of full-parameter fine‑tuning for resource‑constrained space‑qualified hardware, and the underutilization of the visual modality despite its rich semantic content.

To address these challenges, we introduce SCAFusion, a robust multimodal 3D detection framework optimized for lunar surface operations. Our work makes the following key contributions:

\begin{enumerate}
    \item We propose a dedicated Section-aware Coordinate Attention (SCA) module that explicitly enhances feature discrimination and spatial localization for small and irregular objects—a capability of paramount importance for reliable perception in planetary robotics.
    
    \item A parameter‑efficient Cognitive Adapter into the camera backbone, enabling effective adaptation of pre‑trained models with minimal tunable parameters, thus significantly reducing fine‑tuning overhead.
    
    \item A Contrastive Alignment Module (CAM) that operates during the Lift‑Splat‑Shoot projection to enforce consistency between camera‑based semantics and LiDAR‑based geometry, thereby strengthening cross‑modal feature fusion.
    
    \item A Camera Auxiliary Training Branch that amplifies the representational power of the visual stream during training, ensuring that camera data contributes substantively to the multimodal perception system.
\end{enumerate}

We evaluate SCAFusion extensively, both on the terrestrial nuScenes[5] benchmark and—as our primary validation scenario—in a high‑fidelity lunar simulation environment constructed using the NVIDIA Isaac Sim physics engine. Experimental results demonstrate consistent and significant improvements over the BEVFusion[1] baseline, with particularly notable gains in small‑object detection performance, thereby validating the effectiveness of our framework for off‑world perception tasks.

\begin{figure*}[t] 
    \centering
    \includegraphics[width=\textwidth]  {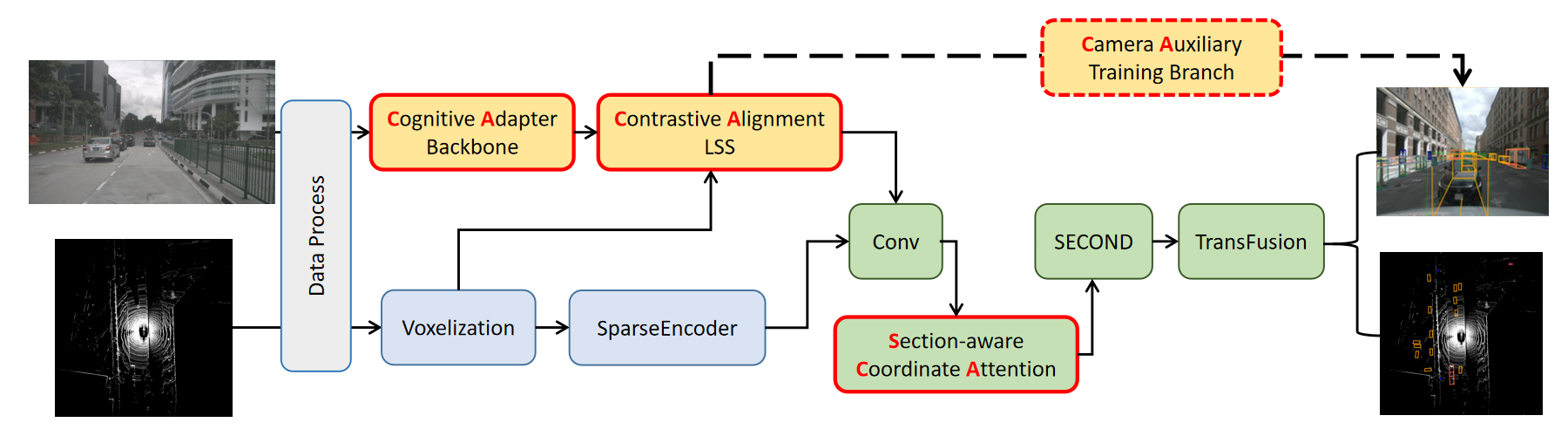}
    \caption{Overall pipeline of SCAFusion. Firstly, the model extracts multi-modal features through specific backbones. Then image features are transformed into image-BEV features after enhancing semantic consistency through Contrastive Alignment Module. Concurrently, image-BEV features fuse with lidar-BEV features and improve the detection capability of small target objects through the Section-aware Coordinate Attention module. The Cognitive Adapter in the camera backbone freezes the pre-training weights of the original camera backbone during backpropagation and adjust fewer parameters to achieve better performance. Only during the training process, image-BEV features are processed through Camera Auxiliary Training Branch(CATB) to fully tap the potential of camera information in multimodal fusion process.}
    \label{SCAFusion_Pipeline}
\end{figure*}

\section{Related Works}
\label{sec:related}

\subsection{Camera-based 3D Perception}

Multi‑camera 3D object detection methods are broadly classified into two categories: BEV‑based methods, which rely on geometric perspective transformation, and Transformer‑based query methods.

BEV‑based approaches[6,8] employs the Lift‑Splat‑Shoot (LSS)[7] technique to project features from perspective view to BEV representation. Further studies[9,10] leverage long‑term cyclic fusion to integrate temporal information continuously.

Query‑based methods[11,12,13,14] decode image features via Transformer architectures, generate target‑specific queries from a 2D detector to strengthen detection. SparseBEV[15] introduces a fully sparse detection framework with scale‑adaptive attention and adaptive spatiotemporal sampling to dynamically capture temporal-BEV features.

\subsection{Lidar-based 3D Perception}

LiDAR is extensively adopted in perception algorithms for its capacity to deliver dense, centimeter‑accurate point cloud data, making it highly suitable for planetary rover navigation. Based on feature extraction strategies, LiDAR‑based 3D perception methods are broadly categorized into two approaches: point‑based methods that directly process raw point clouds, and voxel‑based methods that first convert points into volumetric grids.

Point-based methods[16,17,18] introduces permutation invariance  to handle point cloud disorder while preserving spatial information through max‑pooling. 3DSSD[19] enhances the set abstraction layer by combining feature‑aware and distance‑aware farthest point sampling, boosting both accuracy and inference speed.

Among voxel‑based methods[20,21] partition point clouds into uniform 3D voxels and encode intra‑voxel points into a sparse feature representation, integrating feature extraction and detection into one end‑to‑end network.. However, fixed voxel sizes often struggle to accommodate large scale variations common in natural terrain.

\subsection{Multi-sensor Fusion}

Multimodal fusion methods are commonly categorized by their fusion strategy into three types: (1) primary‑secondary fusion, which designates either image or LiDAR as the dominant modality; (2) BEV‑space fusion, which unifies multi‑sensor features into a shared bird’s‑eye‑view representation; and (3) query‑based fusion, where object queries interact with multimodal features through attention mechanisms.

F‑PointNets[22], however, heavily depends on the performance of the 2D detector. EA‑LSS[23] introduces radar‑guided frustum generation and employs an Edge‑Aware Depth Fusion module to reduce depth discontinuity, along with a Fine‑Grained Depth module to align predicted and actual depth scales.

BEVFusion[1] projects both camera and LiDAR features into a unified BEV space and fuses them via convolution, yet still suffers from cross‑modal misalignment and under‑utilization of visual information. Follow‑up works[24,25,26] have since extended this paradigm to enhance multimodal integration. Nevertheless, these methods often struggle to maintain feature consistency in visually sparse or geometrically irregular environments—conditions characteristic of lunar surfaces.

\section{METHOD}
\label{sec:method}
This chapter will explain the SCAFusion model improved by BEVFusion, and the overall pipeline is shown in Fig. \ref{SCAFusion_Pipeline}.

\subsection{Camera Backbone based on Cognitive Adapter} 

We incorporate the Cognitive Adapter called Mona[28] module within each Swin-Transformer[27] Block and conducted training through incremental parameter tuning, achieving superior performance compared to full parameter tuning. We made adjustments to the original architecture of Swin-Transformer: the last stage of the original Swin-T four stage structure was discarded, and only the first three stages were retained for feature extraction, with each stage containing 2 Swin-T Blocks.The overall structure of the improved camera image processing branch is illustrated in Fig. \ref{Mona_Backbone}:
\begin{figure}[htb] 
    \centering
        \includegraphics[scale=0.3]  {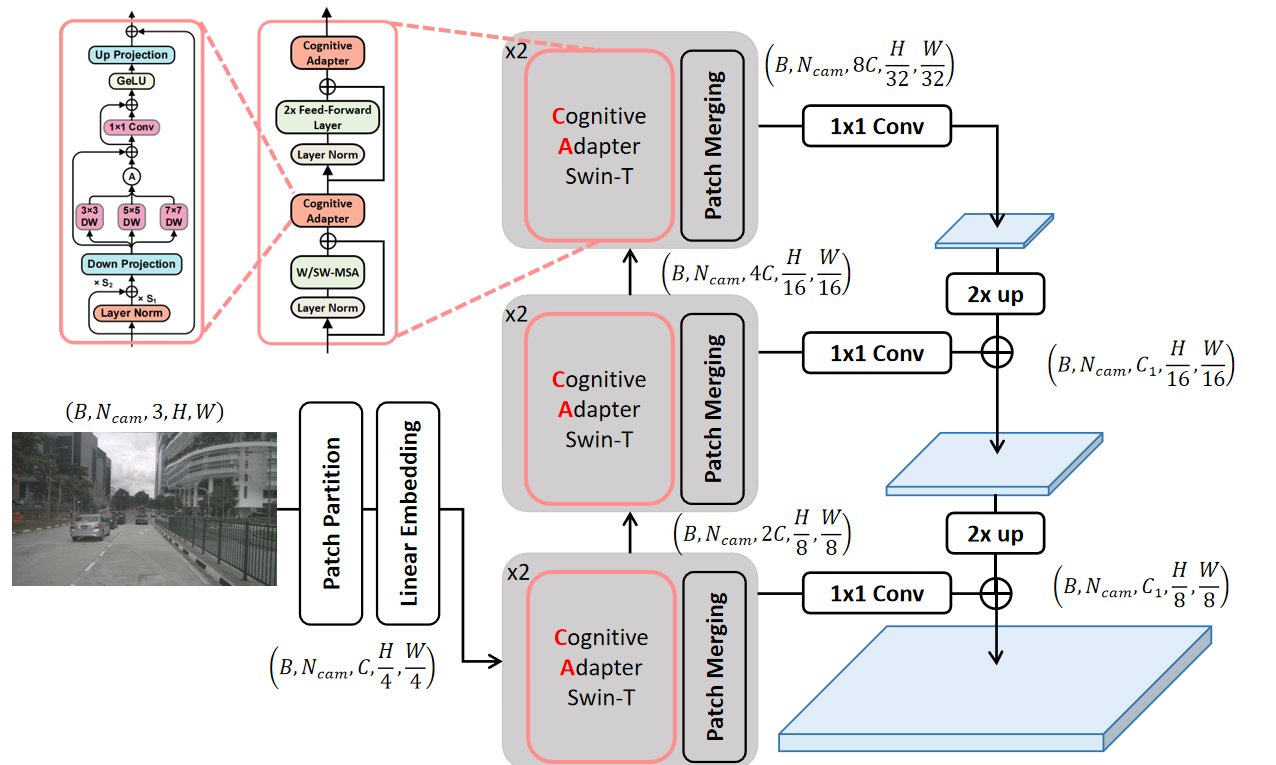}
        \caption{Illustration of \textbf{C}ognitive-\textbf{A}dapter-based backbone.}
        \label{Mona_Backbone}
\end{figure}

For the Cognitive Adapter module itself, it includes modules such as dimensionality reduction, multi cognitive visual filters, activation functions, and dimensional escalation, and incorporates skip connections within the adapter to enhance the model's adaptability. This structural design enables Mona to significantly improve the performance of visual tasks while maintaining high efficiency. It is mainly based on Multi-Cognitive Visual Filters, which utilizes three parallel depth-wise convolutions to capture features within different receptive field ranges; and it combines channel down-projection, up-projection, and multiple residual connections to achieve efficient gradient propagation and parameter adjustment while being lightweight. Given the input feature $x_{img}^i $ of the Swin-T backbone network in stage , the processing process of the Mona module can be represented by the following formula:\\

\begin{equation}
    \resizebox{\columnwidth}{!}{
        $\displaystyle
        \left\{
        \begin{aligned}
            & x_{img}^{i'} = x_{img}^i + \mathbf{U}_i \mathrm{Sigmoid} (\mathrm{Conv}_{1\times1} (\mathrm{Conv}_{dw} \mathbf{D}_i (x_{norm}^i))) \\
         & x_{norm}^i = s_1 \cdot \mathbf{LN} (x_{img}^i) + s_2 \cdot x_{img}^i
        \end{aligned}
        \right.
        $%
}
\end{equation}

where $ \mathbf{LN}(\cdot) $ represents Layer Normalization processing, $ \mathbf{U}(\cdot) $ and $ \mathbf{D}(\cdot) $ respectively represent up-projection and down-projection, $ \mathrm{Conv}_{dw} $ and $ \mathrm{Conv}_{1\times1} $ respectively represent multi-scale depthwise separable convolution with residual connections and ordinary convolution operations.

\subsection{Contrastive Alignment enhancement for Lift-Splat-Shoot}

\begin{figure}[htb] 
    \centering
        \includegraphics[scale=0.25]  {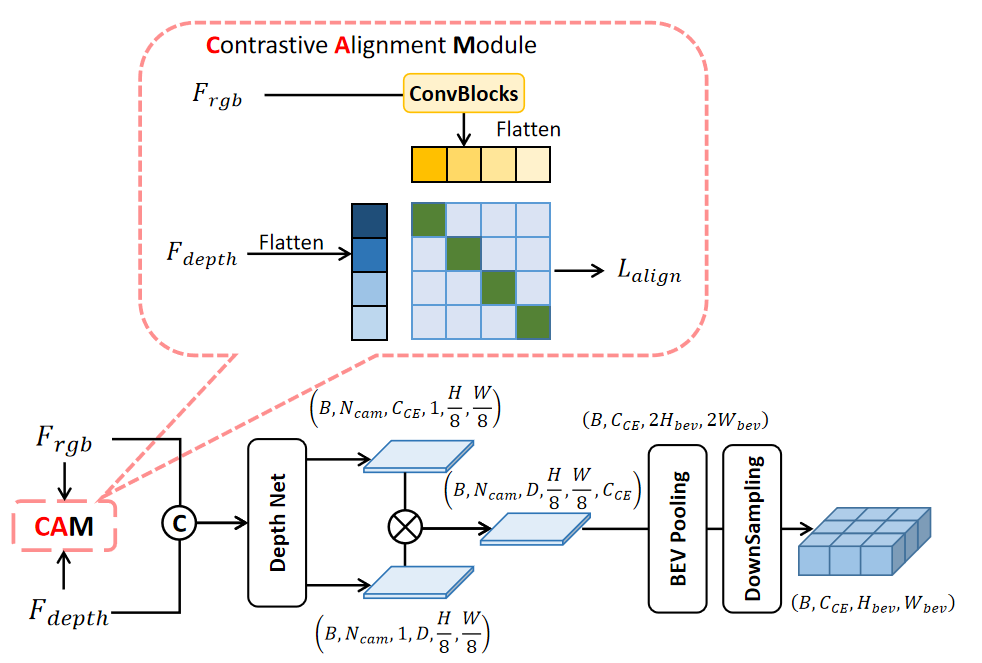}
        \caption{Illustration of Contrastive Alignment Module. RGB features are aligned with depth features through the alignment loss.}
        \label{CA_LSS}
\end{figure}

During the training process, a Contrastive Alignment Module ($\mathbf{CAM}$) is incorporated at the input of LSS to provide additional supervisory signals for the model. This module aligns RGB and depth features and maintains their semantic consistency, as illustrated in Fig. \ref{CA_LSS}. In practical implementation, referring to the Normalized Temperature-Scaled Cross-Entropy Loss (NT-Xent) proposed by Chen T et al. [36], CAM not only enhances the similarity of RGB-depth feature pairs from the same camera perspective of the same sample, but also increases the dissimilarity of RGB-depth feature pairs from different samples or cameras.

Firstly, the RGB and depth features are prepocessed to be expanded into one-dimensional vectors, which have the same length, three convolutional blocks are used to adjust the number of channels layer by layer to make it consistent with $ F_{rgb} $, where $ F_{rgb}^{'} , F_{depth}^{'} \in \mathbf{R}^{B \cdot C_{CE} \cdot H_{bev} \cdot W_{bev}} $ and $ B $ denotes the size of the mini-batch used for training.

Next, based on NT-Xent, we utilize $ F_{rgb}^{'} $ and $ F_{depth}^{'} $ calculate $ \mathcal{L}_{align} $ ,the alignment loss of the model, where the hyperparameter $ \tau $ are used to control the degree of alignment sharpening:\\

\begin{equation}
    \resizebox{\columnwidth}{!}{
    $\displaystyle
        \left\{
        \begin{aligned}
            & \mathcal{L}_{align} = -\frac{1}{BC_{CE}} \sum_{i=1}^{BC_{CE}} \log \frac{\exp(\mathrm{sim}(F_{rgb_i}^{'}, F_{depth_i}^{'}))/\tau}{\sum_{j=1}^{BC_{CE}} \exp(\mathrm{sim}(F_{rgb_i}^{'}, F_{depth_j}^{'}))/\tau} \\
            & \mathrm{sim}(F_{rgb_i}^{'} , F_{depth_j}^{'}) = \frac{F_{rgb_i}^{'} \cdot F_{depth_j}^{'}}{\|F_{rgb_i}^{'}\| \|F_{depth_j}^{'}\|}
        \end{aligned}
        \right.
        $%
    }
\end{equation}

Besides, it should be noted that since $ \mathcal{L}_{align} $ is related to the number of cameras and batch size, when the number of cameras is fixed, the batch size should be appropriately increased during training.

\subsection{Section-aware Coordinate Attention Module for small object detection}

\begin{figure}[htb] 
    \centering
        \includegraphics[scale=0.5]  {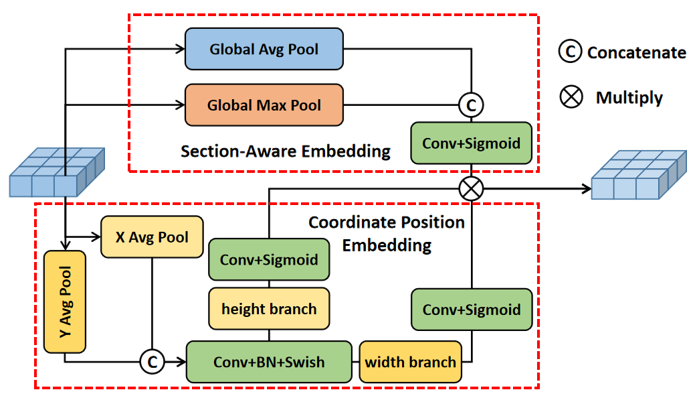}
        \caption{Illustration of Section-aware Coordinate Attention Module.}
        \label{SCA_Attention}
\end{figure}

The Coordinate Attention attention module ingeniously embeds positional information into channel attention by decomposing two-dimensional global pooling into one-dimensional pooling encoding in the X and Y directions. Compared to the CBAM [29] module, it can establish long-range spatial dependencies between channels and has a lower parameter count, however the local importance is ignored. We propose the Section-aware Coordinate Attention Module for small object detection tasks. Its internal structure is shown in Fig. \ref{SCA_Attention}.

The input feature is separately inject into Section-Aware Embedding Module(SAEM) and Coordinate Position Embedding Module(CPEM). The CPEM first performs one-dimensional global average pooling on the input along the horizontal and vertical directions, the generated intermediate features of the c-th channel on two different spatial directions can be formulated as:

\begin{equation}
    \left\{
    \begin{aligned}
        & z_{c}^{h} (h) = \frac{1}{W} \sum_{0 \leq i \leq W} x_{c}(h,i) \\
        & z_{c}^{w} (w) = \frac{1}{H} \sum_{0 \leq j \leq H} x_{c}(j,w)
    \end{aligned}
    \right.
\end{equation}

Then, the features in the two directions are concatenated and subjected to nonlinear transformation through a shared $ 1 \times 1$ convolution. Afterwards, they are split into horizontal and vertical components and activated by sigmoid to generate direction sensitive channel attention weights: 

\begin{equation}
    \left\{
    \begin{aligned}
        & f_{co} = \delta(\mathrm{concat}[z^{h},z^{w}]) \\
        & w^{dire} = \sigma(\mathbf{conv}_{1\times1}(f_{co}^{dire})), \quad dire = h,w 
    \end{aligned}
    \right.
\end{equation}

where $\delta$ is the non-linear activation function and $\sigma$ is the sigmoid function.

The SAEM focuses on important areas in the entire spatial position through Global Average Pool and Global Max Pool, the section-aware weight can be formulated as:

\begin{equation}
    \left\{
    \begin{aligned}
        & z_{c}^{a} (x) = \mathrm{AvgPool}(x) \\
        & z_{c}^{g} (x) = \mathrm{MaxPool}(x) \\
        & f_{se} = \mathrm{concat}[z^a, z^g] \\
        & w^{se} = \sigma(\mathbf{conv}_{1\times1}(f_{se})) 
    \end{aligned}
    \right.
\end{equation}

Finally, the attention maps are dot product processed with the original input to output the features $y_c(i,j)$:

\begin{equation}                                                                                 
y_c(i,j) = x_c(i,j) \times w_c^h(i) \times w_c^w(j) \times w_c^{s}(j).
\end{equation}

\subsection{Camera Auxiliary Training Branch}

The camera auxiliary training branch receives camera BEV features $ x_{CE} $ and first completes phased feature extraction through the ResNet  backbone network, where each $ResBlock(\cdot)$ consists of two residual blocks. In the first and second stages, downsampling operations are performed while adjusting the number of channels, And the third stage only adjusts the number of channels:

\begin{equation}
    \left\{
    \begin{aligned}
        & x_{CE}^{i} = ResBlocks_{i}(x_{CE}^{i-1}),\quad i=1,2,3 \\
        & x_{CE}^{0} = x_{CE}
    \end{aligned}
    \right.
\end{equation}

Next, we use FPN* to fuse the first stage feature $ x_{CE}^{1} \in \mathbf{R}^{\frac{C_{aux}}{2} \times \frac{H_{bev}}{2} \times \frac{W_{bev}}{2}}$ and the third stage feature $ x_{CE}^{3} \in \mathbf{R}^{2C_{aux} \times \frac{H_{bev}}{4} \times \frac{W_{bev}}{4}}$, the input feature $ x_{aux} \in \mathbf{R}^{C_{aux} \times H_{bev} \times W_{bev}} $ of the detection head is obtained, where $ \mathrm{Up}(\cdot) $ uses bilinear interpolation for upsampling: 

\begin{equation}
x_{aux} = \mathrm{Up}( \mathrm{ConvBlock}( Concat( \mathrm{Up}( x_{CE}^{3} ), x_{CE}^{1} ) ) ) .
\end{equation}

The process of feature extraction and FPN* feature fusion in the ResNet backbone network of the camera assisted training branch mentioned above is shown in Fig. \ref{CamAux_Pipeline}:
\begin{figure}[htb] 
    \centering
        \includegraphics[scale=0.6]  {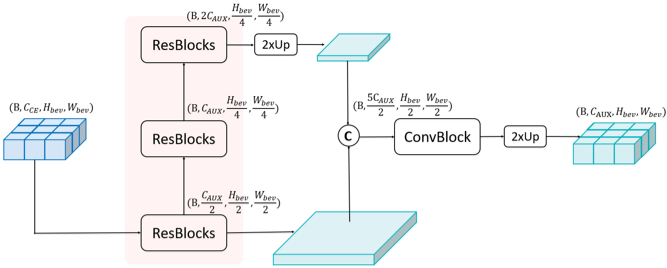}
        \caption{Illustration of Camera Auxiliary Training Branch.}
        \label{CamAux_Pipeline}
\end{figure}

Receive feature $ x_{aux} $ from FPN*, first use convolution to adjust the channel dimension, and obtain feature $ x_{aux}^{0} \in \mathrm{R}^{C_{ctr} \times H_{bev} \times W_{bev}} $. 

Next, various parameters for the 3D detection task are predicted from $ x_{aux}^{0} $, including the center point coordinates, 3D dimensions, direction, and category of the detection box:

\begin{equation}
    \hat{y}_{aux}^{para} = \mathrm{FFN}_{para}( x_{aux}^{0} ), \quad para = c,h,dim,rot,cls \\
\end{equation}

Finally, the original output of the network (without using NMS during training) is restored to the true physical scale and decoded into a standard 3D detection box format to calculate various losses.

\section{Experiments}
\label{sec:experiment}

This chapter will demonstrate the improvement of the model compared to the benchmark on the nuScenes dataset, and conduct experiments on the simulation dataset in the Isaac Sim engine to verify the model's ability to detect small target objects.

\subsection{Primary Evaluation: Lunar Simulation with Isaac Sim}
We construct a photorealistic lunar‑like environment in NVIDIA Isaac Sim, featuring uneven terrain, craters including multiple protrusions and depressions, and two object categories: Meteor (small, irregular) and Platform (larger, regular).

The inspection robot body is equipped with the following sensor configurations: a 32-line LiDAR (10 Hz), a RGB camera with 1900 $\times$ 1200 resolution (10 Hz), and odometer (20 Hz).

\begin{figure}[htb] 
    \centering
        \includegraphics[scale=0.61]  {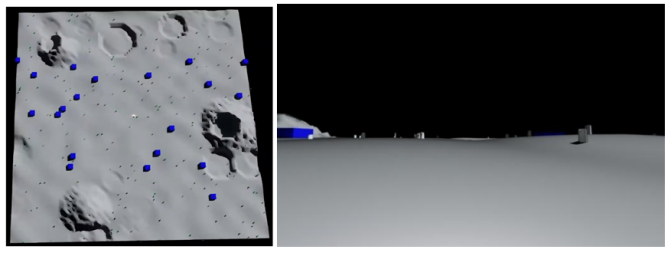}
        \caption{Visualization of simulation environment. Global perspective (left) and front RGB camera perspective (right).}
        \label{isaac_env}
\end{figure}

Considering the impact of lighting changes on model performance, this study collected data under varying illumination, each containing 10 ROS data packets. The duration of a single data packet is about 5 minutes(approximately 3,000 frames), and the data includes topic data such as /odom, /pc\_stcan, /rgf\_data.  Subsequently, using our self-developed Bag2nuS script, we converted the ROS data package into nuScenes format (annotated at 2Hz) and simultaneously modified the source code of the nuscenes devkit library for direct training and testing on the network. The test results of the model are shown in Table \ref{isaac_SOTA}:

\begin{table}[h]
    \centering
    \caption{Test results and AP metrics for each category on Isaac Sim simulation dataset}
    \label{isaac_SOTA}
    \resizebox{\columnwidth}{!}{
        \begin{tabular}{|c|cc|ccc|cc|}
            \hline
            Method & mAP$\uparrow$ & NDS$\uparrow$ & mATE$\downarrow$ & mASE$\downarrow$ & mAOE$\downarrow$ & Meteor$\uparrow$ & Platform$\uparrow$ \\
            \hline
                IS-Fusion      &      71.06     &      66.92     &      0.1054     &      0.0729     &      0.6826         &      74.6      &      67.5     \\
                BEVFusion      &      79.38     &      73.34     &      0.1147     &      0.0801     & 0.4398         &      76.4      &      82.9     \\
                MV2DFusion[33] &      81.46     &      68.50     &      0.1251     &      0.0975     &      1.5453         &      71.7      &      91.2     \\
            \textbf{SCAFusion} & \textbf{90.93} & \textbf{82.68} & \textbf{0.0905} & \textbf{0.0352} &     \textbf{0.1532}    & \textbf{86.8}  & \textbf{95.0} \\
            \hline
        \end{tabular}
    }
\end{table}

\begin{figure}[htb] 
    \centering
        \includegraphics[scale=0.5]  {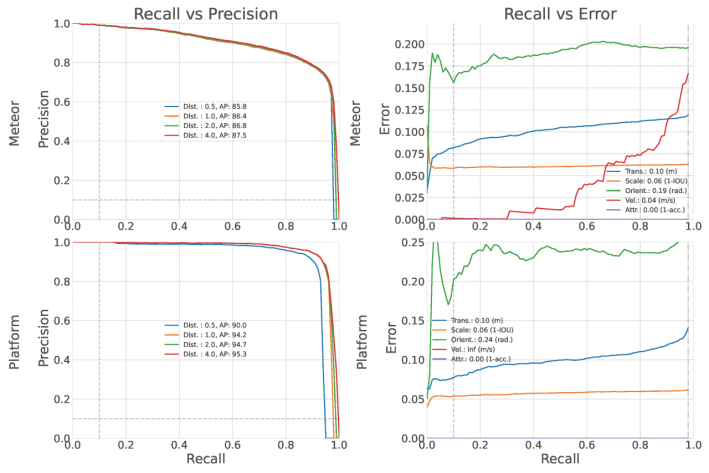}
        \caption{P-R/E-R curves of SCAFusion on simulation datasets.}
        \label{isaac_curves}
\end{figure}

The results show that our proposed SCAFusion model has improved the average accuracy (mAP) index by 11.55\% compared to the BEVFusion benchmark model, reaching 90.93\%, and performs better than MV2DFusion and IS-Fusion, which ranks top and fourth respectively on the nuScenes-det leaderboard. Meanwhile, this study also recorded the accuracy recall curve and error recall curve of SCAFusion on various categories of the simulation dataset, as shown in Fig. \ref{isaac_curves}.

From the comparison of the above curves, it can be observed that although the volume of Meteor is significantly smaller than that of Platform, its detection accuracy and robustness still maintain a high level at different distance thresholds. This may be related to the higher annotation proportion of Meteor in the Isaac Sim datasets, indicating that sufficient data can effectively compensate for the detection challenge caused by the small size of the target object to some extent.

\begin{figure}[htb] 
    \centering
        \includegraphics[scale=0.55]  {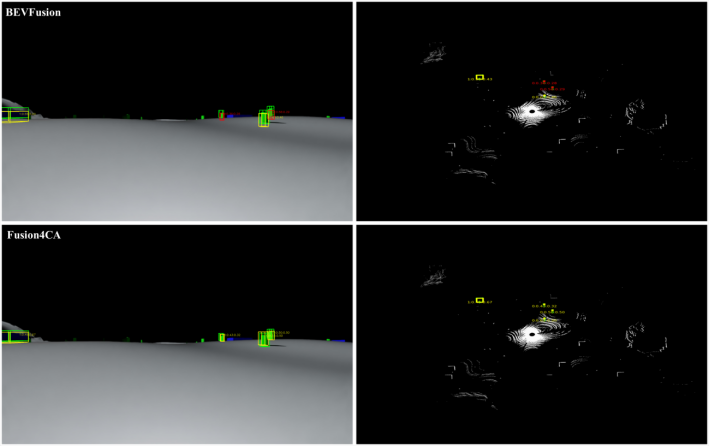}
        \caption{Visualization results of BEVFusion(up) and SCAFusion(down) on the Isaac Sim dataset.}
        \label{isaac_vis}
\end{figure}

Fig. \ref{isaac_vis} shows the visual comparison results of SCAFusion and BEVFusion on the simulation dataset. In this figure, the green bounding box represents the true annotation, the yellow bounding box represents predictions that are correct in both category and position (while meeting category consistency and the set IoU threshold requirements), and the red bounding box identifies prediction results with category errors or positioning deviations. Through this intuitive comparison method, the performance difference between the two algorithms in object detection tasks can be clearly evaluated. It can be seen that compared to the BEVFusion baseline, SCAFusion proposed in this study has fewer red bounding boxes, indicating higher accuracy.

\subsection{Ablation Study}
This section introduces the ablation experiment based on SCAFusion for the nuScenes dataset to demonstrate the effectiveness of the model improvement. The SCAFusion model only adds CA attention module (0.01M) and Mona module (1.41M) compared to the BEVFusion model (40.80M) during inference, and its increased inference overhead (3.48\%) is almost negligible. In addition, it should be noted that due to time costs and limitations on the evaluation of the test set by nuScenes officials, this study only evaluated the ablation experiment on the validation set based on 6 rounds(RTX 4090 and 741000 iterations), and took the model weight corresponding to the highest mAP index round. The test results of the ablation experiment are as follows in Table \ref{nus_Ablation}:

\begin{table}[h]
    \centering
    \caption{Ablation Study}
    \label{nus_Ablation}
    \resizebox{\columnwidth}{!}{
        \begin{tabular}{|c|cccc|cc|}
            \hline
            Order & CAM & CamAux & SCA-Attention & Mona & mAP & NDS \\
            \hline
            01(BEVFusion) & / & / & / & / & 64.7 & 69.4 \\
            02 & $\surd$ & / & / & / & 67.0 & 70.4 \\
            03 & / & $\surd$ & / & / & 68.7 & 71.5 \\
            04 & / & / & $\surd$ & / & 64.6 & 69.4 \\
            \hline
            05 & $\surd$ & $\surd$ & / & / & 68.9 & 71.5 \\
            06 & $\surd$ & $\surd$ & $\surd$ & / & 69.3 & 71.7 \\
            07(SCAFusion) & $\surd$ & $\surd$ & $\surd$ & $\surd$ & \textbf{69.7} & \textbf{72.1} \\
            \hline
        \end{tabular}
    }
\end{table}

\begin{figure}[htb] 
    \centering 
        \includegraphics[scale=0.5]  {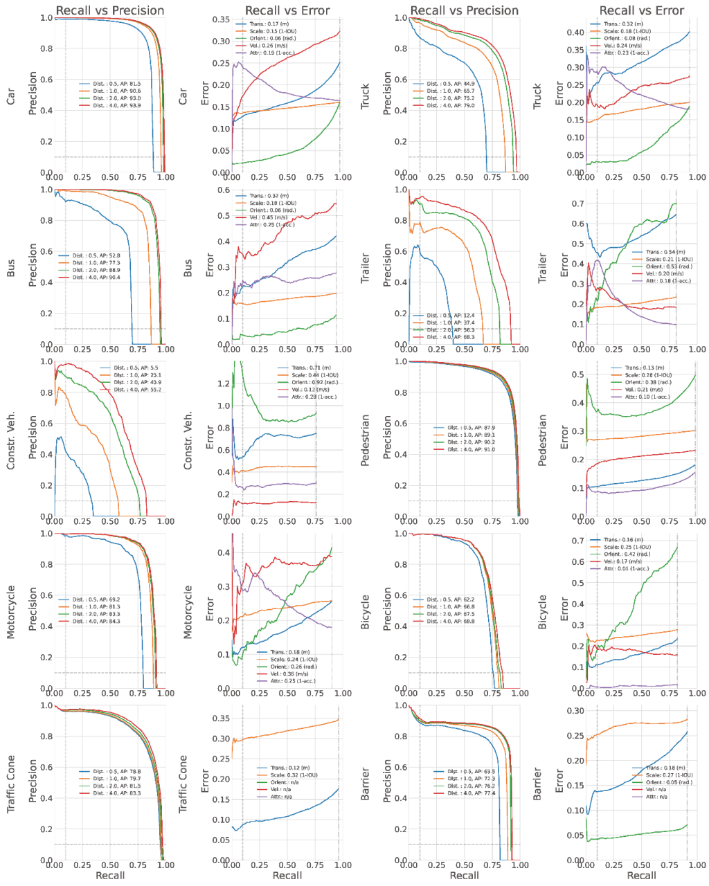}
        \caption{P-R and E-R curves of SCAFusion on nuScenes.}
        \label{nus_curves}
\end{figure}

The results indicate that under the same training configuration, SCAFusion achieved a 5.0\% mAP improvement on the nuScenes validation set compared to the BEVFusion baseline, reaching 69.7\%, while the NDS index reached 72.1\%, an improvement of 2.7\%. This study also attempted to add the Mona module separately to the BEVFusion baseline, but under the same training configuration, unstable training may occur. At the same time, this study also attempted to remove the camera branch based on the BEVFusion baseline and only retain the LiDAR branch for training, in order to explore the actual effect of multimodal fusion. The experimental results show that under the same training configuration, the single-mode model that only retains the LiDAR point cloud branch has 64.6\% mAP and 69.2\% NDS, which are only 0.1\% and 0.2\% lower than the corresponding indicators of the BEVFusion baseline, respectively. This indicates that the BEVFusion baseline has not fully explored the information of camera modalities in multimodal fusion, thus further demonstrating the effectiveness of many improvements in this study. In addition, this study recorded the accuracy recall curve and error recall curve of SCAFusion on various categories of the nuScenes dataset, as shown in Fig. \ref{nus_curves}.

It can be seen that for the Construction Vehicle and Trailer classes, their accuracy changes sharply with the setting of distance thresholds, and even drops to 5.5\% and 12.4\% in the strictest cases, becoming the weakness of mAP. This study speculates that this may be related to their low annotation levels (1.26\%, 2.13\%) on the nuScenes dataset.

\subsection{Comparison on the nuScenes dataset}

\begin{figure}[htb] 
    \centering
        \includegraphics[scale=0.5]  {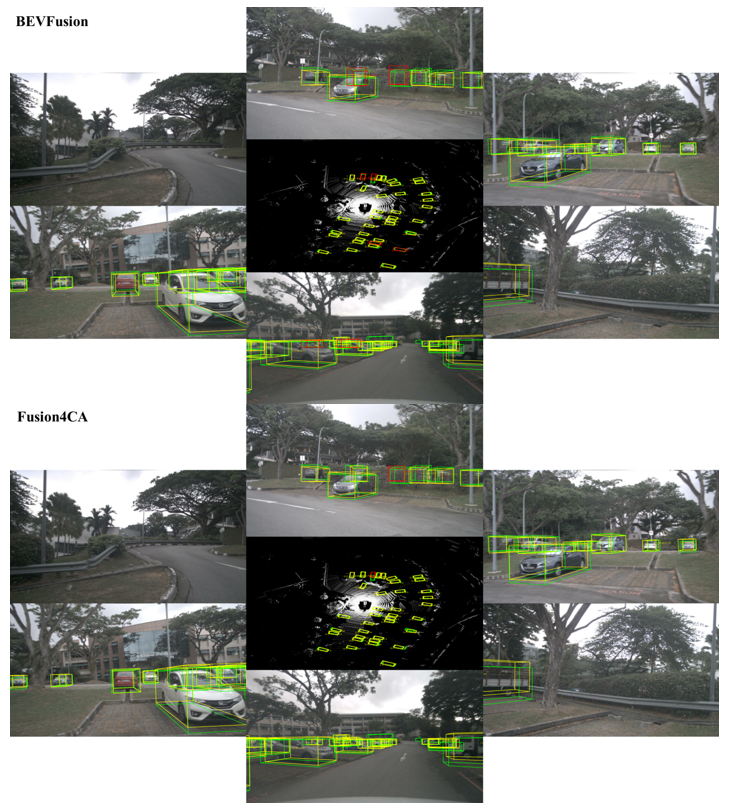}
        \caption{Visualization results of BEVFusion(up) and SCAFusion(down) on nuScenes dataset.}
        \label{nus_vis}
\end{figure}

The visualization results between SCAFusion and BEVFusion on the nuScenes dataset are shown in Fig. \ref{nus_vis}. It can be seen that compared to the baseline, the SCAFusion model proposed in this study has fewer red bounding boxes (less misdetections), indicating higher accuracy.

This study conducted comparative experiments between the SCAFusion model and previous state-of-the-art methods on the nuScenes Test Set, and the results are shown in Table \ref{nus_SOTA}. The experimental results show that the SCAFusion model achieved 69.7\% mAP and 71.9\% NDS without using Test time Augmentation (TTA) after only 6 rounds of training, and surpassed existing advanced methods in multiple key indicators, demonstrating its superior performance.

\begin{table*}[t]
    \centering
    \caption{Comparison between other algorithms}
    \label{nus_SOTA}
        \begin{tabularx}{\textwidth}{@{}lcc*{10}{>{\centering\arraybackslash}X}@{}}  
            \hline
            Method & mAP & NDS & Car & Truck & Construction Vehicle & Bus & Trailer & Barrier & Motorcycle & Bicycle & Pedestrian & Traffic Cone \\
            \hline
            PointPillar[34] & 30.5 & 45.3 & 68.4 & 23.0 & 4.1 & 28.2 & 23.4 & 38.9 & 27.4 & 1.1 & 59.4 & 30.8 \\
            CneterPoint[30] & 60.3 & 67.3 & 85.2 & 53.5 & 20.0 & 63.6 & 56.0 & 71.1 & 59.5 & 30.7 & 84.6 & 78.4 \\
            TransFusion-L[31] & 65.5 & 70.2 & 86.2 & 56.7 & 28.2 & 66.3 & 58.8 & 78.2 & 68.3 & 44.2 & 86.1 & 82.0 \\
            \hline
            PointPainting[35] & 46.4 & 58.1 & 77.9 & 35.8 & 15.8 & 36.2 & 37.3 & 60.2 & 41.5 & 24.1 & 73.3 & 62.4 \\
            MVP[37] & 66.4 & 70.5 & 86.8 & 58.5 & 26.1 & 67.4 & 57.3 & 74.8 & 70.0 & 49.3 & 89.1 & 85.0 \\
            GraphAlign[37] & 66.5 & 70.6 & 87.6 & 57.7 & 26.1 & 66.2 & 57.8 & 74.1 & 72.5 & 49.0 & 87.2 & 86.3 \\
            FusionPainting[38] & 68.1 & 71.6 & 87.1 & 60.8 & 30.0 & 68.5 & 61.7 & 71.8 & \textbf{74.7} & \textbf{53.5} & 88.3 & 85.0 \\
            TransFusion[31] & 68.9 & 71.7 & 87.1 & 60.0 & 33.1 & 68.3 & 60.8 & 78.1 & 73.6 & 52.9 & 88.4 & \textbf{86.7} \\
            BEVFusion-PKU[2] & 69.2 & 71.8 & 88.1 & 60.9 & 34.4 & 69.3 & 62.1 & \textbf{78.2} & 72.2 & 52.2 & 89.2 & 85.2 \\
            \textbf{SCAFusion(Ours)} & \textbf{69.7} & \textbf{71.9} & \textbf{88.7} & \textbf{61.4} & \textbf{36.6} & \textbf{72.4} & \textbf{63.5} & 74.5 & 74.3 & 50.1 & \textbf{89.3} & 86.4 \\
            \hline
        \end{tabularx}
\end{table*}


\section{Conclusions}
\label{sec:conclusion}

We present SCAFusion, a multimodal 3D detection framework designed for lunar surface exploration. By integrating a Section-aware Coordinate Attention module specifically aimed at small‑object detection, together with lightweight adapters and alignment mechanisms, the model significantly improves detection accuracy in standard autonomous driving benchmarks and achieve SOTA performance on our task-based simulation datasets. 

Although SCAFusion performs less effectively on nuScenes dataset, our model still offers a practical, efficient, and robust perception solution for future planetary robotics missions where reliable detection of small, irregular obstacles is paramount. The improvement of detection ability for small target objects was specifically verified on the Isaac Sim simulation dataset, with a 11.55\% increase and a 9.47\% increase in mAP index compared to BEVFusion and MV2DFusion, respectively.

Our future work will be dedicated to improving the generalization and robustness of the model to achieve state-of-the-art (SOTA) performance even on public datasets.


\end{document}